\pdfoutput=1
\documentclass{article}
\usepackage{iclr2026_conference,times}
\iclrfinalcopy

\usepackage[utf8]{inputenc}
\usepackage[T1]{fontenc}
\usepackage{CJKutf8}
\usepackage{amsmath,amssymb,amsfonts}
\usepackage{booktabs}
\usepackage{longtable}
\usepackage{graphicx}
\usepackage{array}
\usepackage{float}
\usepackage{placeins}
\usepackage{multirow}
\usepackage{enumitem}
\usepackage{fvextra}
\usepackage{url}
\usepackage{hyperref}
\usepackage{xspace}
\usepackage{tikz}
\usetikzlibrary{arrows.meta,positioning,fit,shapes.misc}
\hypersetup{hidelinks}
\makeatletter
\def\fps@figure{t!}
\def\fps@table{t!}
\makeatother
\newcommand{\method}{BayesPO\xspace}
\DefineVerbatimEnvironment{CodeBlock}{Verbatim}{fontsize=\small,breaklines=true,breakanywhere=true,breaksymbolleft={},breaksymbolright={}}

\newcommand{\emb}{\theta}

\title{\method: Bayesian Prompt Optimization via Parallel-Tempered Gradient-Guided Discrete MCMC}

\author{Junjie Zhou \quad Zhijian Ou\thanks{Corresponding author (ozj@tsinghua.edu.cn). This work is partly supported by the National Science and Technology Major Project (2023ZD0121401).}\\
Speech Processing and Machine Intelligence (SPMI) Lab\\
Tsinghua University, China}

\begin{document}
\begin{CJK*}{UTF8}{gbsn}
\maketitle

\begin{abstract}
Prompt optimization adapts large language models (LLMs) without updating model parameters, but many automatic prompt optimizers remain heuristic search procedures over candidate instructions. This paper studies prompt optimization as Bayesian posterior sampling over discrete prompt tokens. We define a posterior distribution by combining a task likelihood term, which rewards prompts that explain input-output examples, with a language-model prior, which favors fluent instructions. This converts prompt optimization into an energy-based posterior sampling problem, for which gradients can be used to guide discrete Markov chain Monte Carlo (MCMC) proposals over vocabulary tokens. We refer to our framework as \method, short for Bayesian Prompt Optimization. In this paper, \method is instantiated with Markov chain Monte Carlo: it uses a Metropolis-Hastings corrected Gibbs-with-Langevin (GwL) proposal and integrates parallel tempering for global exploration of rugged LLM-induced energy landscapes. The concrete sampler further adapts the GwL sampler to the practical constraints of non-weight-tied LLM embeddings. Experiments with Qwen2.5 models show that the sampler discovers semantically meaningful prompts on diagnostic tasks, that parallel tempering helps escape a local optimum in a poetry completion task, and that post-optimizing APE prompts on 24 instruction-induction subtasks improves average accuracy from 60.04\% to 63.23\%. The study also reveals two main limitations: energy minimization may overfit small optimization sets, and the current sampler remains computationally expensive. These findings position Bayesian prompt sampling as a principled post-optimization tool and point to a promising direction for probabilistic prompt optimization.
\end{abstract}

\section{Introduction}

Large language models (LLMs) can be adapted to many downstream tasks by modifying only the natural-language prompt. This property makes prompt optimization attractive when updating model parameters is unnecessary, costly, or undesirable. However, prompt optimization is a hard combinatorial search problem. A prompt is a sequence of discrete tokens, so the number of possible prompts grows exponentially with length; moreover, evaluating one candidate prompt usually requires at least one LLM forward pass.

Existing automatic prompt optimization methods follow several major paradigms. Gradient-based methods such as AutoPrompt use token-level gradients to greedily replace prompt tokens \citep{shin2020autoprompt}. LLM-generation methods such as Automatic Prompt Engineering (APE) sample candidate task descriptions from input-output examples and select the best candidates by downstream evaluation \citep{zhou2023large}. More recent methods use natural-language feedback or textual gradients to edit prompts iteratively \citep{pryzant2023automatic,das2024greater}. These approaches are effective in many cases, but they are often designed as heuristic optimization procedures. They optimize a finite candidate set, perform local greedy edits, or rely on an LLM to generate the next candidate. They do not explicitly define a posterior distribution over prompts, nor do they provide a Markov chain whose stationary distribution is the desired prompt posterior.

This paper asks whether prompt optimization can be formulated as posterior sampling in a discrete token space. We treat the prompt as a discrete random variable conditioned on a small set of input-output examples. The posterior combines two signals: a task likelihood that measures whether a prompt induces the desired outputs, and a language-model prior that encourages the prompt itself to remain natural. This posterior can be written as an energy-based model (EBM) \citep{lecun2006tutorial,ou2024energy}, making it possible to use MCMC methods. The central challenge is that random-walk MCMC is inefficient in a high-dimensional discrete vocabulary, while continuous relaxations introduce a mismatch when projected back to tokens.

\begin{figure}[t!]
\centering
\resizebox{\textwidth}{!}{%
\begin{tikzpicture}[
  node distance=0.65cm and 0.75cm,
  box/.style={draw, rounded corners, align=center, minimum height=0.9cm, minimum width=2.4cm, fill=gray!8},
  smallbox/.style={draw, rounded corners, align=center, minimum height=0.75cm, minimum width=2.1cm, fill=gray!8},
  arrow/.style={-Latex, thick}
]
\node[box] (data) {Few-shot\\input-output pairs};
\node[box, right=of data] (energy) {Prompt posterior\\$P(\rho\mid \mathcal{D}) \propto e^{-U(\rho)}$};
\node[box, right=of energy] (grad) {Energy gradient\\$\nabla U(\emb)$};
\node[box, right=of grad] (proposal) {Discrete proposal\\over vocabulary};
\node[box, right=of proposal] (mh) {MH correction\\accept / reject};
\node[smallbox, below=of grad] (pt) {Parallel tempering\\multi-temperature chains};
\node[box, right=of mh] (out) {Best target-chain\\prompt};
\draw[arrow] (data) -- (energy);
\draw[arrow] (energy) -- (grad);
\draw[arrow] (grad) -- (proposal);
\draw[arrow] (proposal) -- (mh);
\draw[arrow] (mh) -- (out);
\draw[arrow] (pt) -- (proposal);
\draw[arrow] (mh.south) |- (pt.east);
\end{tikzpicture}}
\caption{Overview of \method. A prompt posterior over discrete token sequences is converted into an energy. A gradient-guided vocabulary proposal and MH correction produce a valid discrete MCMC update. Parallel tempering runs several temperature-scaled chains and swaps states to improve global exploration.}
\label{fig:overview}
\end{figure}
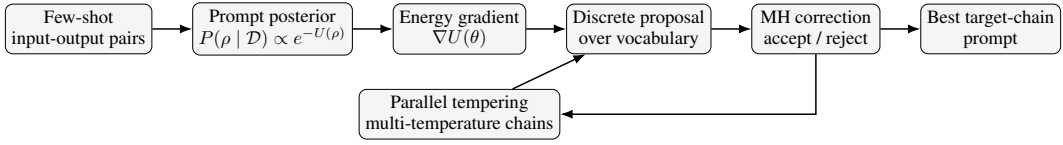

We therefore propose \method, a Bayesian prompt optimization framework based on parallel-tempered gradient-guided discrete MCMC. Figure~\ref{fig:overview} summarizes the method. We first construct a prompt energy from task likelihood and prompt prior. We then compute the gradient of this energy with respect to prompt embeddings and use it to build a discrete proposal over vocabulary tokens. A Metropolis-Hastings (MH) correction preserves the target distribution. In practice, full-sequence gradient-guided discrete proposals \citep{du2024principled} often have very low acceptance rates, because changing many tokens simultaneously can sharply increase energy. We therefore use a Gibbs-with-Langevin (GwL) variant \citep{du2024principled} that updates one prompt position at a time and immediately applies accepted changes. Finally, because LLM-induced energy landscapes can be multimodal, we use parallel tempering (PT) \citep{hukushima1996exchange,geyer2011importance}: high-temperature chains explore broadly, while swap moves transfer promising states back to the target-temperature chain.

Our experiments provide both positive evidence and cautionary observations. On two small semantic transformation tasks, the sampler decreases energy and discovers prompts that are aligned with the tasks. In a poetry completion task, a single GwL chain remains trapped near an incorrect completion, while PT transfers a better state to the target chain and recovers the correct line. On the APE instruction induction benchmark, starting from APE prompts, \method improves average test accuracy over 24 subtasks by 3.19 percentage points. The largest gains occur when the APE prompt is initially semantically misaligned. However, the sampler can also degrade accuracy even as the training energy decreases, revealing an energy-accuracy mismatch under extremely small training sets.

The contributions of this work are:
\begin{itemize}[leftmargin=1.3em]
  \item We formulate prompt optimization as Bayesian posterior sampling over discrete prompt tokens, with an energy function combining task likelihood and an LLM prompt prior.
  \item We adapt gradient-guided discrete MCMC to prompt optimization through an MH-corrected GwL sampler, including a treatment of non-weight-tied input and output embeddings.
  \item We integrate parallel tempering to improve global exploration, showing its value in a local-optimum poetry completion task and using it as a key exploration component in the APE instruction-induction benchmark.
  \item We conduct an empirical study on the APE instruction induction benchmark, analyzing gains, failures, computation, and the mismatch between lower training energy and test accuracy.
\end{itemize}

\section{Related Work}

\paragraph{Automatic prompt optimization.}
AutoPrompt searches for discrete trigger tokens using gradients from masked language models \citep{shin2020autoprompt}. APE treats prompt optimization as black-box candidate generation and evaluation, showing that LLMs can infer task instructions from examples \citep{zhou2023large}. ProTeGi edits prompts using natural-language gradients and beam search \citep{pryzant2023automatic}. GReaTer extends gradient-based prompt optimization to reasoning traces \citep{das2024greater}. Surveys of prompt optimization emphasize the diversity of candidate generation, evaluation, and search operators \citep{cui2025survey}. \method is complementary to these methods: it can take a generated or manually written prompt as initialization, and then refine it under an explicit posterior energy with MCMC updates.

\paragraph{Bayesian optimization versus Bayesian sampling for prompts.}
Several recent works also use the word ``Bayesian'' in prompt optimization, but the underlying formulation differs from ours. Tomar et al. propose BODE-GEN for test-driven code generation, where Bayesian Optimization (BO) searches for prompts by fitting surrogate models in a continuous embedding space and using an auxiliary LLM to bridge discrete prompts and continuous representations \citep{tomar2025bodegen}. Sabbatella et al. similarly formulate prompt optimization as a black-box combinatorial optimization problem and apply BO over a continuous relaxation of the prompt space \citep{sabbatella2025bayesian}. These methods solve an optimization problem with BO: the goal is to adaptively select the next candidate prompt to evaluate. In contrast, \method formulates prompt optimization as a Monte Carlo sampling problem: we define an energy-based posterior distribution over discrete prompt tokens and build an MH-corrected Markov chain to sample from it. This distinction matters because our proposal is not a Gaussian-process surrogate acquisition loop; it is a gradient-guided MCMC transition kernel with a target stationary distribution.

\paragraph{Black-box prompt optimization and BPO.}
Black-Box Prompt Optimization (BPO) optimizes user prompts for LLM alignment without model training and leverages human preferences to improve win rates \citep{cheng2024black}. We deliberately avoid using the bare acronym BPO for our method because BPO already strongly refers to Black-Box Prompt Optimization. More importantly, the technical setting is different: BPO is a black-box alignment-oriented prompt optimizer driven by preference feedback, whereas \method is a white-box or gray-box Bayesian sampler that uses gradients of an explicit prompt energy, MH correction, and parallel tempering. The methods are therefore orthogonal: black-box preference optimization and posterior sampling address different prompt optimization regimes.

\paragraph{Gibbs sampling for prompt inference.}
Reprompting uses Gibbs sampling to infer chain-of-thought (CoT) recipes for reasoning tasks \citep{xu2024reprompting}. It iteratively samples new CoT recipes using previously sampled recipes as parent prompts and an LLM as the proposal mechanism. Reprompting is close in spirit to our work because both use sampling ideas for prompt inference. The differences are threefold. First, Reprompting targets CoT recipes, while \method optimizes general instruction prompts. Second, Reprompting samples natural-language recipes with LLM-generated proposals, while \method constructs token-level discrete proposals from energy gradients. Third, \method includes an explicit EBM posterior, MH correction, and PT, whereas Reprompting is not designed as gradient-guided posterior sampling over vocabulary tokens.

\paragraph{Energy-based models and discrete MCMC for text.}
Energy-based models define unnormalized distributions through scalar energies \citep{lecun2006tutorial}. EBMs have been used and studied broadly in speech and language processing, including language modeling, sequence labeling, speech recognition, and text generation \citep{ou2024energy}. COLD decoding and MuCoLa optimize continuous relaxations of token embeddings with Langevin dynamics or gradient descent for constrained generation \citep{qin2022cold,kumar2022gradient}. Continuous relaxations are convenient, but the projection back to discrete tokens can bias the distribution. Recent gradient-guided discrete MCMC samplers for text, including the Gibbs-with-Langevin (GwL) variant, construct proposals directly in a finite discrete vocabulary and apply MH correction \citep{du2024principled}. \method adapts this line of work to prompt optimization, where the sampled text is not the final response but the instruction that controls downstream behavior.

\paragraph{Parallel tempering.}
Parallel tempering, also known as exchange Monte Carlo, runs multiple chains at different temperatures and swaps states between chains to improve mixing in multimodal landscapes \citep{hukushima1996exchange,geyer2011importance}. The target-temperature chain preserves the desired distribution, while high-temperature chains cross energy barriers more easily. We use PT to address local modes in LLM-induced prompt energy landscapes.

\section{BayesPO}

This section presents the complete BayesPO sampler. Section~3.1 formulates prompt optimization as posterior sampling over discrete prompt tokens. Section~3.2 defines the task-likelihood and prompt-prior energy. Section~3.3 constructs a gradient-guided discrete proposal and explains its GwL implementation. Section~3.4 adapts the proposal to non-weight-tied LLMs. Section~3.5 adds parallel tempering for global exploration.

\subsection{Problem Formulation}

Let $\rho=(\rho_1,\ldots,\rho_N)$ be a prompt token sequence of fixed length $N$, and let $\mathcal{D}=\{(x_i,y_i)\}_{i=1}^{K}$ be a small optimization set for the task. We model the prompt posterior as
\begin{equation}
P(\rho\mid\mathcal{D}) \propto P(\rho)\left[\prod_{i=1}^{K} P(y_i\mid \rho,x_i)\right],
\label{eq:posterior}
\end{equation}
where $P(\rho)$ is a prompt prior induced by the same or another language model, and $P(y_i\mid \rho,x_i)$ is the LLM likelihood of generating target output $y_i$ when the prompt $\rho$ is used with input $x_i$. Taking the negative log posterior yields the energy
\begin{equation}
U(\rho)=-\log P(\rho)-\sum_{i=1}^{K}\log P(y_i\mid \rho,x_i).
\label{eq:energy}
\end{equation}
The resulting energy-based model defines
\begin{equation}
\pi(\rho) \propto \exp[-U(\rho)].
\end{equation}
Low-energy prompts have high posterior probability, because they both explain the task examples and remain plausible prompt strings under the language model prior.

\subsection{Prompt Energy}

For an instruction-tuned LLM with a chat template, the likelihood term in Eq.~\eqref{eq:energy} is computed by placing the prompt in the global instruction field, the task input in the user field, and the target output in the assistant field. Our experiments instantiate this generic construction with the Qwen-Instruct template; Appendix~\ref{app:chat-template} gives the exact formatting used in all likelihood computations. The negative log-likelihood of the assistant tokens is summed over the $K$ examples. The prompt prior is autoregressive:
\begin{equation}
-\log P(\rho)= -\log P(\rho_1)-\sum_{t=2}^{N}\log P(\rho_t\mid \rho_1,\ldots,\rho_{t-1}).
\label{eq:prior}
\end{equation}
The prior discourages degenerate token sequences and biases the sampler toward fluent instructions. In practice, the strength of the prior can be tuned, but the experiments reported here use the direct sum in Eq.~\eqref{eq:energy}.

\subsection{Gradient-Guided Discrete Proposal}
Let $\theta=(\theta_1,\ldots,\theta_N)\in E^N$ be the embedding sequence corresponding to the discrete prompt $\rho$, and let $E=\{e_v:v\in\mathcal{V}\}$ denote the vocabulary embedding table. With abuse of notation, we write $U(\theta)$ for the same energy $U(\rho)$ evaluated through the LLM computation graph on the embedding sequence induced by $\rho$. The proposal for the discrete transition $q(\rho'\mid\rho)$ is expressed as a proposal $q(\theta'\mid\theta)$ over $E^N$ with $\theta'\in E^N$, as follows.

The full-sequence $\ell_p$-Norm Constrained Gradient (p-NCG) proposal \citep{du2024principled} has the form
\begin{equation}
q(\emb'\mid\emb) \propto \exp\left[-\frac{1}{2}\nabla U(\emb)^\top(\emb'-\emb)-\frac{1}{2\alpha}\|\emb'-\emb\|_p^p\right],
\label{eq:pncg-full}
\end{equation}
where $\alpha$ controls the proposal step size. Because the $\ell_p$ norm term factorizes across token positions, this proposal can be written as
\begin{equation}
q(\emb'\mid\emb)=\prod_{t=1}^{N}q(\emb'_t\mid\emb),
\end{equation}
with
\begin{equation}
q(\emb'_t\mid\emb) \propto \exp\left[-\frac{1}{2}\nabla_t U(\emb)^\top(\emb'_t-\emb_t)-\frac{1}{2\alpha}\|\emb'_t-\emb_t\|_p^p\right],\quad \emb'_t\in E.
\label{eq:pncg}
\end{equation}
A sampled embedding is mapped back to its corresponding vocabulary token to obtain the discrete candidate prompt. This means that the proposal treats word positions as conditionally independent given the current sequence, allowing each word embedding to be sampled in parallel from a vocabulary-sized categorical distribution. The first term favors candidates in a local energy-decreasing direction, while the second term, namely the $\ell_p$ norm term, penalizes excessively large jumps in embedding space.

A full p-NCG proposal updates all positions in parallel, but this can lead to low acceptance rates for prompt optimization because changing many tokens at once produces large energy fluctuations. We therefore use a Gibbs-with-Langevin (GwL) variant \citep{du2024principled}. Each scan randomly permutes the prompt positions; for each position, Eq.~\eqref{eq:pncg} is used to sample a single-token replacement while all other positions remain fixed. The proposal excludes self-transitions because MH rejection already allows the chain to stay at the current state. The accepted replacement is applied immediately, so later positions in the same scan use the updated prompt.

Given a candidate state $\rho'$, the MH acceptance probability is
\begin{equation}
\alpha_{\mathrm{acc}}=\min\left\{1,\frac{\pi(\rho')q(\rho\mid\rho')}{\pi(\rho)q(\rho'\mid\rho)}\right\}.
\label{eq:mh}
\end{equation}
This correction compensates for the asymmetry of the gradient-guided proposal and preserves the target distribution in the ideal chain.

\subsection{Non-Weight-Tied Embeddings}
Many modern LLMs, including Qwen2.5-Instruct models \citep{qwen2025qwen25}, do not tie their input embedding matrix and output projection matrix. This requires care because a prompt token participates in the energy in two ways. In the likelihood term, prompt tokens act as input prefix tokens and contribute gradients through the input embeddings. In the prior term, a prompt token can be a predicted token through the output projection and a prefix token for later positions through the input embeddings. This issue does not arise in the original p-NCG text-generation experiments using GPT-2, because GPT-2 uses tied input and output embeddings \citep{du2024principled}.

Let $[\emb]_i$ and $[\emb]_o$ denote the input and output embeddings associated with the same token sequence. We decompose
\begin{equation}
U(\rho)=U_{\mathrm{like}}([\emb]_i)+U_{\mathrm{prior}}([\emb]_i,[\emb]_o).
\end{equation}
The gradients used in the proposal are
\begin{equation}
\nabla_{[\emb]_i}U=\nabla_{[\emb]_i}U_{\mathrm{like}}+\nabla_{[\emb]_i}U_{\mathrm{prior}},\quad
\nabla_{[\emb]_o}U=\nabla_{[\emb]_o}U_{\mathrm{prior}}.
\end{equation}
Accordingly, the gradient-guided proposal's linear term is modified to combine input-side and output-side gradient contributions:
\begin{equation}
q(\emb'\mid\emb) \propto \exp\left[-\frac{1}{2}\left(\nabla_{[\emb]_i}U^\top([\emb']_i-[\emb]_i)+\nabla_{[\emb]_o}U^\top([\emb']_o-[\emb]_o)\right)-\frac{1}{2\alpha}\|[\emb']_i-[\emb]_i\|_p^p\right].
\label{eq:non-tied-proposal}
\end{equation}
The locality penalty is computed in the input embedding space because this space directly determines the prompt representation seen by the model. This separation makes the sampler applicable to non-weight-tied instruction-tuned LLMs while preserving a discrete vocabulary proposal.

\subsection{Parallel Tempering}

To improve global exploration, \method uses parallel tempering \citep{hukushima1996exchange,geyer2011importance} and runs $M+1$ chains with temperatures $\{T_0,\ldots,T_M\}$, where $T_0=1$ is the target chain. Chain $m$ targets
\begin{equation}
\pi_m(\rho) \propto \exp[-U(\rho)/T_m].
\end{equation}
High-temperature chains flatten the energy landscape and cross barriers more easily. After within-chain GwL scans, adjacent chains attempt state swaps. For chains $i$ and $j$ with inverse temperatures $\beta_i=1/T_i$ and $\beta_j=1/T_j$, the swap acceptance probability is
\begin{equation}
\alpha_{\mathrm{swap}}=\min\{1,\exp[(\beta_i-\beta_j)(U(\rho_i)-U(\rho_j))]\}.
\label{eq:swap}
\end{equation}
The target chain records the lowest-energy prompt encountered during sampling.

\begin{figure}[t!]
\setlength{\fboxsep}{6pt}
\centering
\fbox{\begin{minipage}{0.95\textwidth}
\small
\textbf{Algorithm 1: \method}\vspace{0.3em}

\textbf{Input:} initial prompts $\{\rho^{(m,0)}\}_{m=0}^{M}$, temperatures $\{T_m\}_{m=0}^{M}$, optimization set $\mathcal{D}$, step size $\alpha$, total iterations $S$.\vspace{0.2em}

\textbf{For} $s=1,\ldots,S$:\vspace{-0.4em}
\begin{enumerate}[leftmargin=1.4em,itemsep=0.15em]
\item For each chain $m$, randomly permute the $N$ prompt positions.
\item For each position $t$, compute $\nabla_t U(\rho^{(m)})$, build the discrete vocabulary proposal in Eq.~\eqref{eq:pncg}, sample a candidate replacement, and accept it with Eq.~\eqref{eq:mh} using the temperature-scaled energy $U/T_m$.
\item Sample a neighboring pair of temperatures and attempt a state swap with Eq.~\eqref{eq:swap}.
\item Update the best prompt recorded by the target chain $T_0=1$.
\end{enumerate}\vspace{-0.2em}
\textbf{Output:} the lowest-energy prompt observed on the target chain.
\end{minipage}}
\caption{The complete sampler combines local GwL token updates and parallel-tempered state swaps.}
\label{fig:algorithm}
\end{figure}

\section{Experiments}

This section evaluates BayesPO from simple to realistic settings. Section~4.1 describes the models, implementation, and inference template. Section~4.2 validates whether GwL discovers meaningful prompts on two diagnostic tasks. Section~4.3 isolates the role of PT on a poetry completion task. Sections~4.4 and 4.5 evaluate post-optimization on the APE instruction-induction benchmark and analyze energy-accuracy mismatch. Section~4.6 discusses computational cost.

\subsection{Experimental Setup}

We use Qwen2.5-Instruct models. Diagnostic prompt optimization uses Qwen2.5-0.5B-Instruct to verify basic sampler behavior on small semantic transformation tasks. The poetry completion and APE benchmark experiments use Qwen2.5-7B-Instruct, where the energy landscape is more rugged and PT becomes important. All experiments use the official Qwen-Instruct chat template; Appendix~\ref{app:model-diagnostic} collects the model-template and diagnostic-task details; Appendix~\ref{app:chat-template} specifies the exact system/user/assistant formatting used to compute likelihood terms. Experiments are implemented in PyTorch with mixed precision; generation uses vLLM when applicable. The hardware environment uses a multi-GPU server with NVIDIA RTX 3090 GPUs. Additional experiment details, reproducibility information, and the language-model-use statement are provided first in Appendix~\ref{app:experiment-details}.

\subsection{Diagnostic Prompt Optimization}

We first evaluate whether the sampler can discover task-relevant prompts from random initializations. In both diagnostic tasks, we use $K=4$ input-output examples; the exact task examples $\{(x_i,y_i)\}$ are shown in Appendix~\ref{app:diagnostic-examples}, and the diagnostic-task hyperparameters are reported in Appendix~\ref{app:diagnostic-hparams}. The \emph{classical Chinese translation} task maps English sentences into classical Chinese-style Chinese. The \emph{antonym generation} task maps a Chinese word to its antonym. Because this experiment is intended as a qualitative demonstration and Qwen2.5-0.5B-Instruct has a relatively simple energy landscape, we use single-chain GwL rather than PT and select the best result across ten random seeds.

\begin{figure}[t!]
\centering
\begin{minipage}{0.48\textwidth}
\centering
\includegraphics[width=\linewidth]{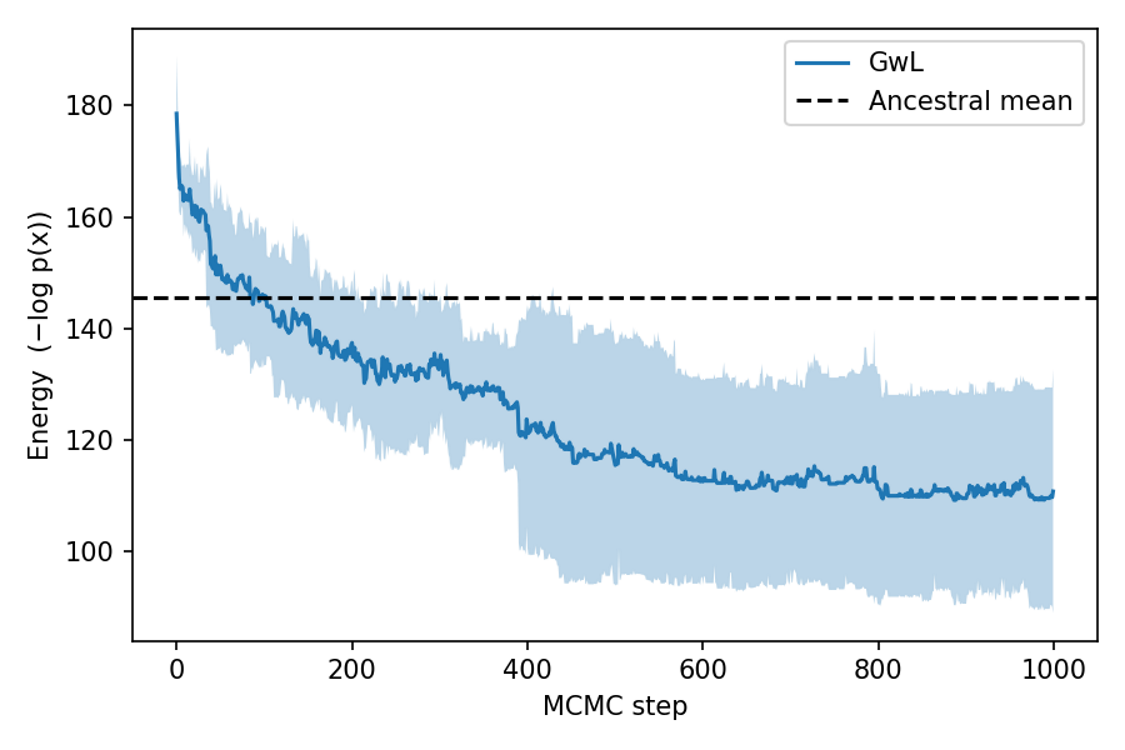}
\end{minipage}\hfill
\begin{minipage}{0.48\textwidth}
\centering
\includegraphics[width=\linewidth]{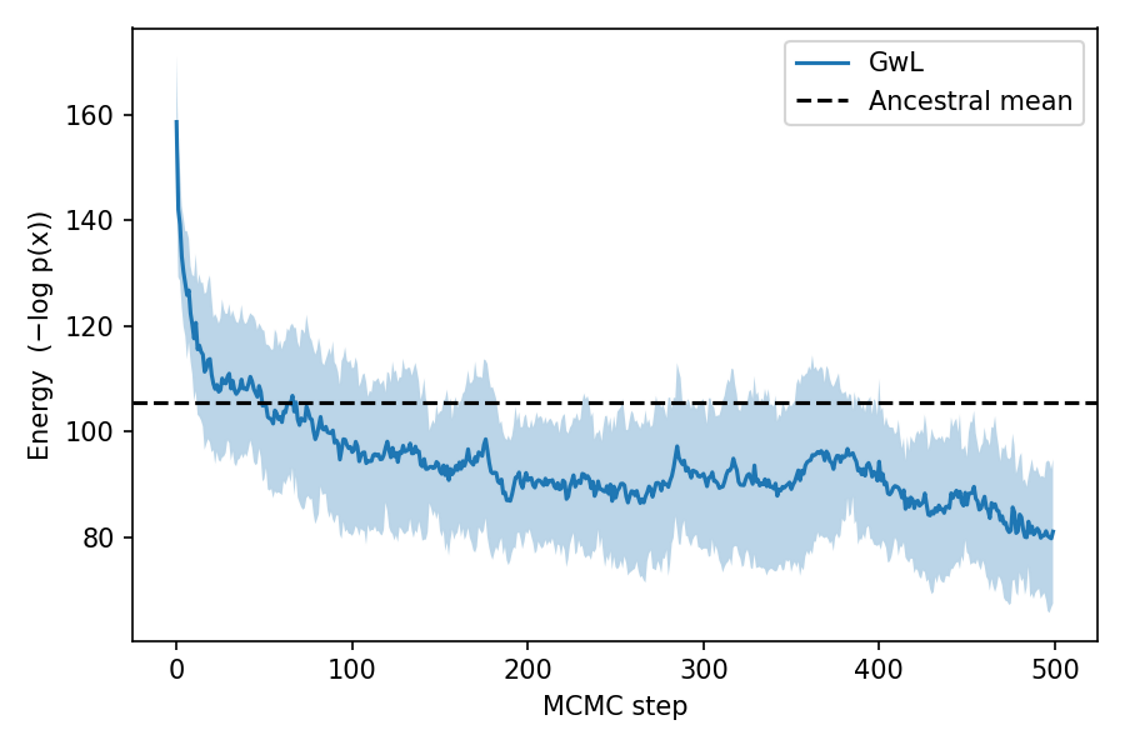}
\end{minipage}
\caption{Diagnostic prompt optimization. Left: classical Chinese translation. Right: antonym generation. The sampler reduces energy and reaches lower-energy prompt regions than ancestral random prompt samples.}
\label{fig:diagnostic}
\end{figure}

After passing the four input-output pairs to the sampler, we run $S=1000$ iterations for the classical Chinese translation task and $S=500$ iterations for the antonym generation task. Figure~\ref{fig:diagnostic} plots the energy trajectories of the target chain during sampling. The dashed line is the mean energy of 100 prior-sampled sequences of fixed length $N=10$ for the classical Chinese task and fixed length $N=20$ for the antonym task. The solid line is the mean trajectory over ten random seeds, and the shaded region denotes one standard deviation over the ten seeds. Energy steadily decreases from the high-energy initial region and converges to lower-energy regions, indicating that the sampler continuously finds prompt sequences that better match the task objective.

For the classical Chinese translation task, the lowest-energy prompt sequence found by the sampler is:
\begin{quote}\emph{编写成一句简单的古代汉字表达式} (English: ``Write it as a simple ancient Chinese expression.'')\end{quote}
For the antonym generation task, the lowest-energy prompt sequence is:
\begin{quote}\emph{生成一个否定的陈述。你可以用任何你想要的词汇} (English: ``Generate a negative statement. You may use any vocabulary you want.'')\end{quote}
The two exact sequences contain task-relevant semantic cues, such as ancient Chinese expression and negation, showing that gradient-guided discrete MCMC can move random prompt strings toward meaningful prompt regions.

\subsection{Parallel Tempering for Poetry Completion}

We next test whether PT helps escape local optima through a poetry completion task based on Li Bai's 《静夜思》 (\emph{Quiet Night Thought}).\footnote{The full poem is: ``床前明月光，疑是地上霜。举头望明月，低头思故乡。'' A literal English rendering is: ``Bright moonlight before my bed; I suspect it is frost on the ground. I raise my head and gaze at the bright moon; I lower my head and think of home.'' This makes the task difficult. Under a prefix such as ``下面是《静夜思》中的两句诗：'', a left-to-right completion is likely to fill the first line, ``床前明月光，'', which corresponds to a local optimum. However, the suffix is the fourth line, ``低头思故乡。''. Considering both the prefix and the suffix, the completion ``举头望明月，'' attains lower energy and acts as the global optimum in this controlled task, because it both follows the prefix and connects naturally to the suffix.} The experiment fixes the prefix as ``下面是《静夜思》中的两句诗：'' (``Below are two lines from \emph{Quiet Night Thought}:'') and the suffix as ``低头思故乡。'' (``I lower my head and think of home.''). The missing middle sequence has a fixed length of six tokens and is initialized as ``床前明月光，'' (``Moonlight shines before my bed,''). This initialization is obtained by giving the prefix and suffix to Qwen2.5-7B-Instruct and letting the model autonomously complete the missing text.\footnote{The template for using Qwen2.5-7B-Instruct to generate the missing line for initialization is shown in Appendix~\ref{app:poetry-init-template}.} The task requires the sampler, guided by the energy function, to generate a middle text that is semantically coherent with the prefix and suffix and consistent with the original poem.

The single-chain GwL baseline uses the same prefix, suffix, initialization, and sampling length, but does not use PT. The PT run uses nine chains with temperatures $\{2.0,1.6,1.33,1.28,1.22,1.17,1.11,1.06,1.0\}$, with $T_0=1.0$ as the target chain, $S=4000$ iterations, and step size $\alpha=4.0$.

\begin{figure}[t!]
\centering
\begin{minipage}{0.49\textwidth}
\centering
\includegraphics[width=\linewidth]{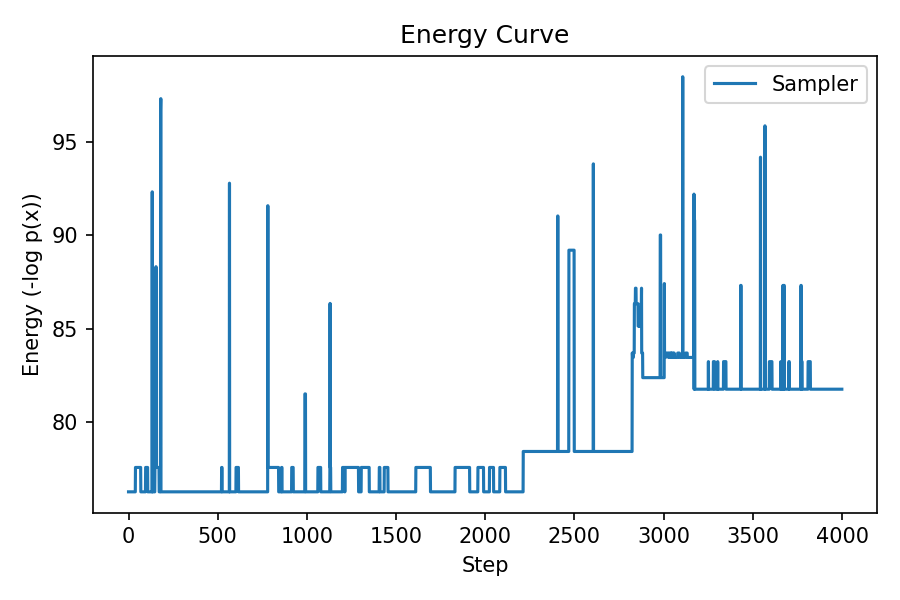}
\end{minipage}\hfill
\begin{minipage}{0.49\textwidth}
\centering
\includegraphics[width=\linewidth]{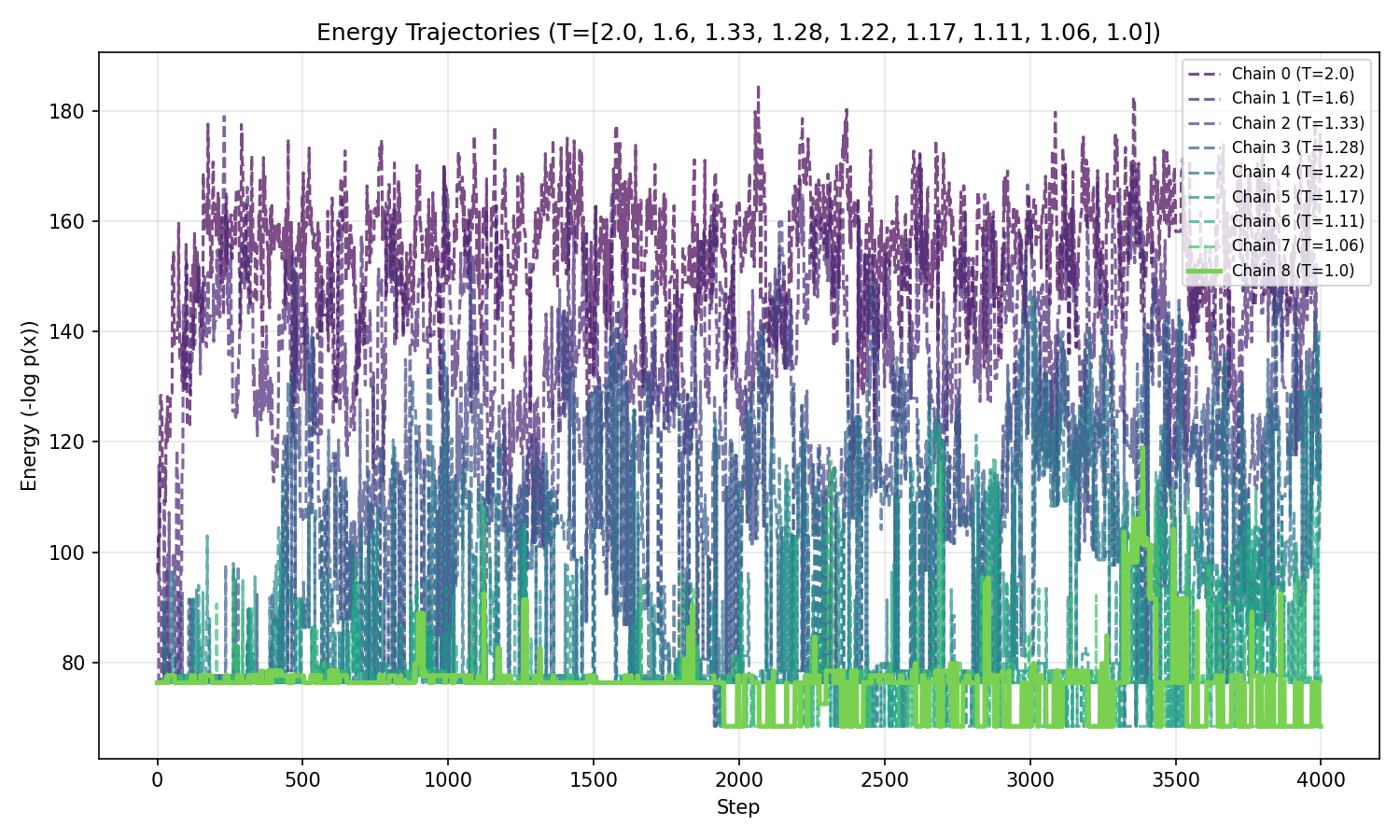}
\end{minipage}
\vspace{0.5em}
\begin{minipage}{0.95\textwidth}
\centering
\includegraphics[width=\linewidth]{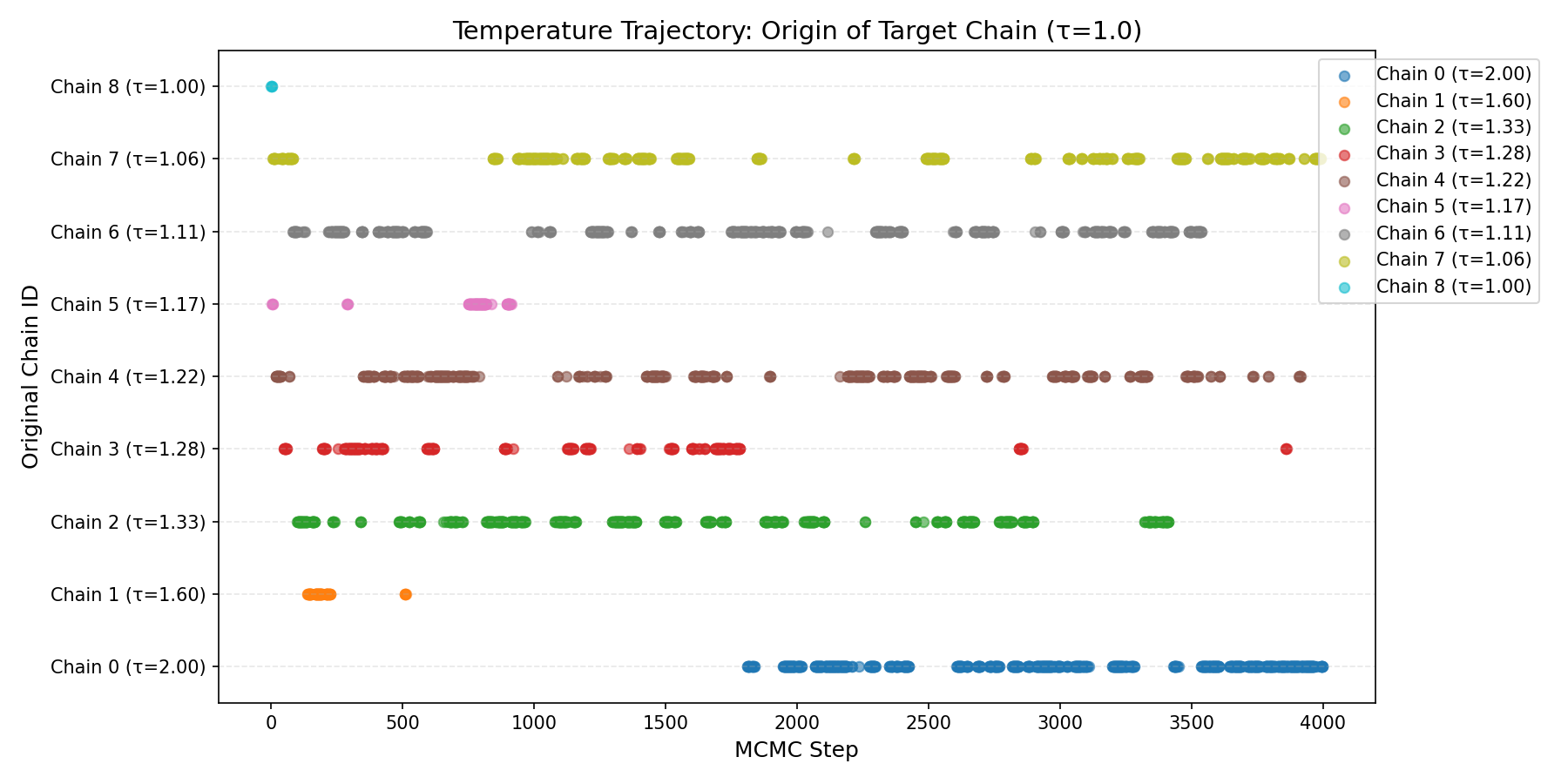}
\end{minipage}
\caption{Parallel tempering for poetry completion. Top-left: a single GwL chain remains trapped. Top-right: PT explores multiple energy regions. Bottom: temperature trajectory of states reaching the target chain, showing extensive state exchange across temperatures.}
\label{fig:pt_poetry}
\end{figure}

As shown in Figure~\ref{fig:pt_poetry} (top-left), the single-chain GwL baseline remains near the initialization and fails to cross the energy barrier; its lowest-energy completion remains ``床前明月光，'' (``Moonlight shines before my bed,''). In contrast, Figure~\ref{fig:pt_poetry} (top-right) shows that PT explores multiple energy regions, and the target chain recovers the correct missing line ``举头望明月，'' (``I raise my head and gaze at the bright moon,''). Figure~\ref{fig:pt_poetry} (bottom) is the temperature traversal plot during parallel-tempered sampling: each dot indicates which original chain supplied the state currently occupying the target chain at that step. The plot shows frequent state exchanges across temperatures. In particular, the high-temperature chain ($T=2.0$) explores broadly and can pass low-energy states back to the target chain, confirming that the improvement is due to multi-temperature exploration rather than local single-chain refinement.

\subsection{APE Instruction Induction Benchmark}

We evaluate on the 24-task instruction induction benchmark used by APE \citep{zhou2023large,honovich2022instruction}. For each task, we randomly select six training examples. Three examples are used for APE candidate generation, three for APE online evaluation, and all six examples are used to construct our energy. The original APE test sets are used for final evaluation. APE prompts serve both as the baseline and as initialization for \method.

We run eight PT chains at temperatures $\{1.8,1.5,1.4,1.33,1.28,1.2,1.1,1.0\}$ for 1000 iterations with $\alpha=4.0$. The optimized prompt length is 20 tokens for all tasks except Sentence Similarity, where it is 50 tokens. A fixed prefix ``The instruction was to'' is used to bias the generated sequence toward an instruction.

\begin{figure}[t!]
\centering
\includegraphics[width=\textwidth]{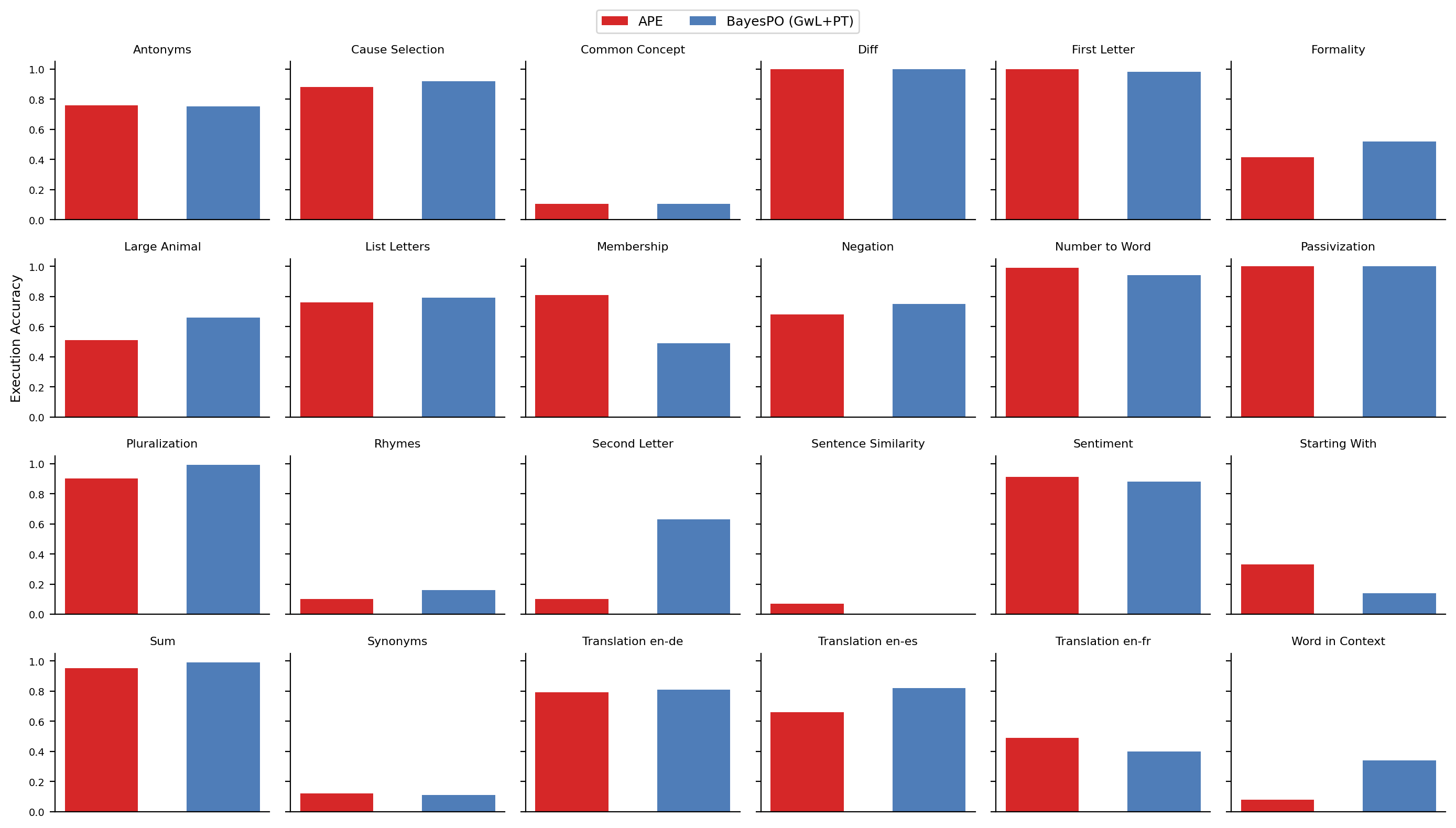}
\caption{Accuracy comparison on all 24 APE instruction induction subtasks. Red bars are APE-initialized prompts, and blue bars are prompts after BayesPO (GwL+PT optimization).}
\label{fig:ape_bar}
\end{figure}

Figure~\ref{fig:ape_bar} and Table~\ref{tab:ape} report the results. The average accuracy improves from 60.04\% to 63.23\%. The largest gains occur on tasks where the APE prompt is initially poor: Second Letter improves from 10\% to 63\%, Word in Context from 8\% to 34\%, Larger Animal from 51\% to 66\%, and English-Spanish translation from 66\% to 82\%. These cases suggest that \method is particularly useful for post-correcting a generated prompt whose task description has semantic drift or weak output-format constraints. Appendix~\ref{app:full-prompts} lists the full before/after prompt pairs and accuracies for all 24 APE subtasks.

\begin{table}[t!]
\centering
\small
\resizebox{\textwidth}{!}{%
\begin{tabular}{lrrr@{\quad}lrrr}
\toprule
Task & APE & BayesPO & $\Delta$ & Task & APE & BayesPO & $\Delta$ \\
\midrule
Passivization & 100.00 & 100.00 & 0.00 & Antonyms & 76.00 & 75.00 & -1.00 \\
Cause Selection & 88.00 & 92.00 & +4.00 & Common Concept & 10.42 & 10.54 & +0.12 \\
Difference & 100.00 & 100.00 & 0.00 & First Letter & 100.00 & 98.00 & -2.00 \\
Formality & 41.56 & 51.96 & +10.40 & Number to Word & 99.00 & 94.00 & -5.00 \\
Larger Animal & 51.00 & 66.00 & +15.00 & List Letters & 76.00 & 79.00 & +3.00 \\
Negation & 68.00 & 75.00 & +7.00 & Starting With & 33.00 & 14.00 & -19.00 \\
Rhymes & 10.00 & 16.00 & +6.00 & Second Letter & 10.00 & 63.00 & +53.00 \\
Sentiment & 91.00 & 88.00 & -3.00 & Pluralization & 90.00 & 99.00 & +9.00 \\
Sentence Similarity & 7.00 & 0.00 & -7.00 & Sum & 95.00 & 99.00 & +4.00 \\
English-German & 79.00 & 81.00 & +2.00 & Synonyms & 12.00 & 11.00 & -1.00 \\
Membership & 81.00 & 49.00 & -32.00 & English-Spanish & 66.00 & 82.00 & +16.00 \\
English-French & 49.00 & 40.00 & -9.00 & Word in Context & 8.00 & 34.00 & +26.00 \\
\midrule
Average & 60.04 & 63.23 & +3.19 & & & & \\
\bottomrule
\end{tabular}}
\caption{Test accuracy on the APE instruction induction benchmark. BayesPO uses the APE prompt as initialization and applies GwL+PT optimization.}
\label{tab:ape}
\end{table}

\subsection{Energy-Accuracy Mismatch}

Table~\ref{tab:energy} highlights a key empirical finding: reducing training energy does not always improve test accuracy. In Second Letter and Word in Context, large energy reductions coincide with large accuracy gains. In Second Letter, the APE prompt drifts toward extracting the penultimate letter, whereas the optimized prompt refocuses on generating the second letter. In Word in Context, the optimized prompt more explicitly constrains the output format by asking the model to type ``same.''

However, Membership and Starting With degrade despite lower energy. Membership drops from 81\% to 49\%; its optimized prompt introduces an overly specific length constraint that likely fits the six optimization examples but fails on the full test distribution. Starting With similarly changes a general ``input sentence'' instruction into a narrower ``selected SMS'' framing. These failures indicate overfitting to spurious patterns in extremely small optimization sets.

\begin{table}[t!]
\centering
\small
\begin{tabular}{lrrrr}
\toprule
Task & $U_{\mathrm{APE}}$ & $U_{\mathrm{BayesPO}}$ & Acc. APE & Acc. BayesPO \\
\midrule
Second Letter & 178.19 & 91.04 & 10.00 & 63.00 \\
Word in Context & 236.09 & 129.09 & 8.00 & 34.00 \\
Membership & 324.07 & 147.27 & 81.00 & 49.00 \\
Starting With & 190.53 & 141.22 & 33.00 & 14.00 \\
Difference & 85.57 & 72.44 & 100.00 & 100.00 \\
Passivization & 109.56 & 66.18 & 100.00 & 100.00 \\
\bottomrule
\end{tabular}
\caption{Representative energy and accuracy changes. Energy decreases in all examples, but generalization can either improve or degrade.}
\label{tab:energy}
\end{table}

\begin{figure}[t!]
\centering
\begin{minipage}{0.49\textwidth}
\centering
\includegraphics[width=\linewidth]{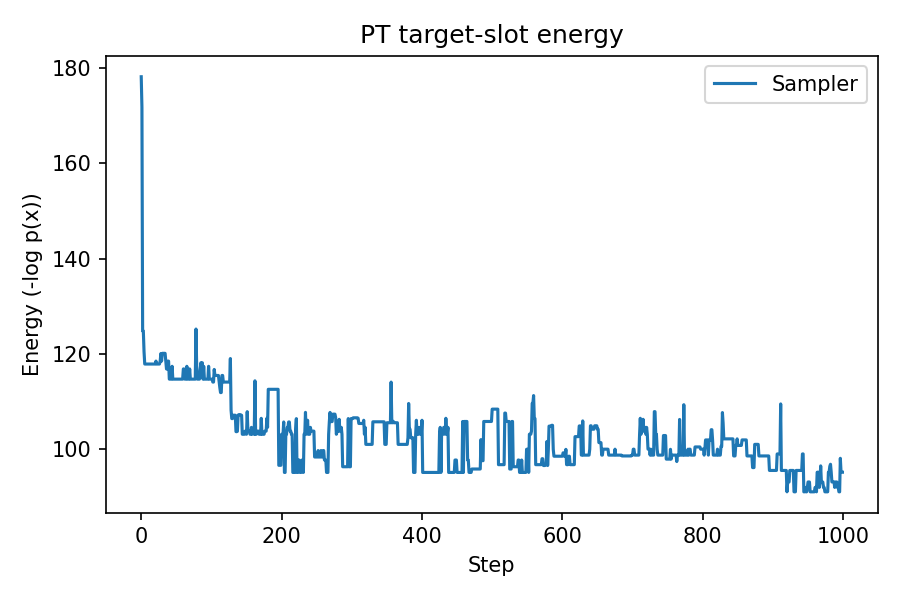}
\end{minipage}\hfill
\begin{minipage}{0.49\textwidth}
\centering
\includegraphics[width=\linewidth]{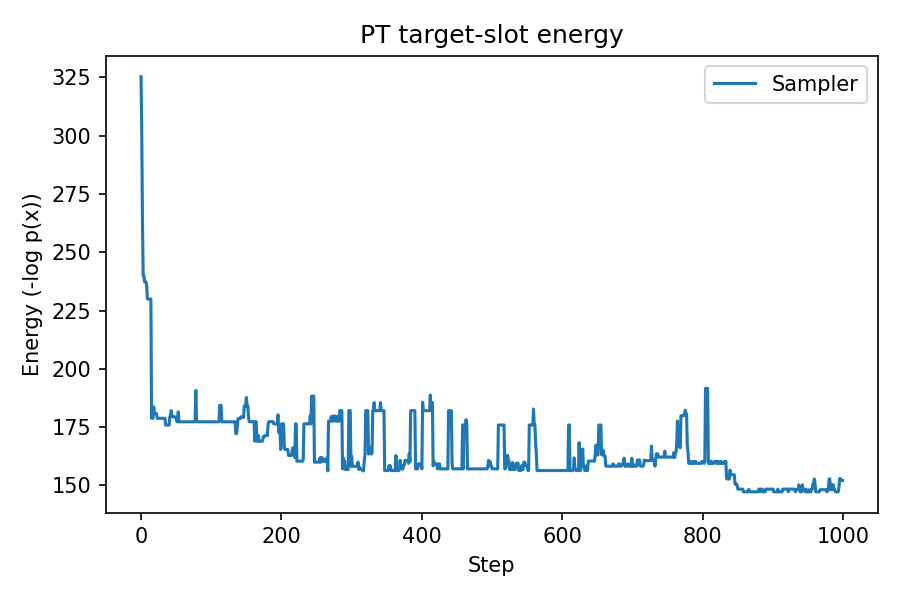}
\end{minipage}
\caption{Representative APE-task energy trajectories. Left: Second Letter, where energy reduction improves accuracy. Right: Membership, where energy reduction overfits and accuracy drops.}
\label{fig:energy_examples}
\end{figure}

Figure~\ref{fig:energy_examples} shows representative target-chain energy trajectories. Additional trajectories for improved, degraded, and stable tasks are provided in Appendix~\ref{app:ape-energy-trajectories}. The decreasing curves confirm that the sampler optimizes the specified training energy. The contrasting accuracy outcomes show that the energy itself must be designed or validated carefully when only a few examples are available.

\subsection{Computational Cost}

Let $N$ be prompt length, $S$ be the number of global iterations, and $M+1$ be the number of temperature chains. A GwL scan updates $N$ positions, and each position requires forward/backward computation for the current state and a forward computation for MH evaluation. Thus the wall-clock cost scales approximately as
\begin{equation}
\mathcal{T}_{\mathrm{GwL+PT}} = \Theta(SMN C_{\mathrm{model}}),
\end{equation}
where $C_{\mathrm{model}}$ includes forward, backward, and MH-evaluation costs. By contrast, APE requires forward-only candidate generation and evaluation. In the current implementation, the sampler is roughly one order of magnitude slower than APE on a single optimization run. This cost is acceptable only when prompt optimization can be amortized over repeated deployment, when prompts must be highly reliable, or when a strong initialization still contains local semantic errors.

\section{Discussion}

\textbf{First, computational cost and Monte Carlo efficiency.}
BayesPO is currently computation-intensive because it uses many MC iterations, repeated backward passes, and multiple temperature chains. One direct direction is to design faster-mixing Monte Carlo methods for high-dimensional discrete prompt spaces. Another possible direction concerns the interaction between BayesPO and the LLM pretraining objective. Current autoregressive LLMs are trained with next-token prediction, which is essentially suffix prediction from a left context. In contrast, Bayesian posterior sampling for prompt optimization behaves more like a prefix-prediction problem: the sampler must infer a good preceding instruction from desired downstream input-output behavior. These two objectives are not fully aligned. A speculative but interesting possibility is that LLM training could incorporate prefix-prediction or instruction-inversion objectives, so that the resulting model provides more informative gradients and better proposal distributions for BayesPO, thereby improving MC efficiency.

\textbf{Second, fixed prompt length.}
The current sampler assumes a fixed prompt length, whereas natural instructions often benefit from insertions, deletions, and variable-length edits. Trans-dimensional Monte Carlo methods offer a promising path beyond this restriction. In particular, trans-dimensional random fields were developed to model observations of different lengths in language modeling \citep{wang2018transdimensional}. Future BayesPO variants could combine token replacement moves with birth-death or reversible-jump style moves so that the Markov chain explores prompt strings of different lengths while maintaining a well-defined target distribution.

\textbf{Third, small optimization sets and energy-accuracy mismatch.}
The APE experiments show that small optimization sets can cause overfitting: a lower training energy may correspond to a prompt that fits a few examples but generalizes worse. When the number of task demonstration examples increases, mini-batch stochastic-gradient sampling becomes attractive. Stochastic Gradient Langevin Dynamics (SGLD) provides a continuous-space Bayesian sampling method with convergence guarantees under annealed step sizes \citep{welling2011sgld}. For discrete prompt optimization, however, gradient-guided MCMC still needs MH correction to preserve the target distribution. Discrete-space mini-batch MCMC with valid or approximately valid correction is therefore an important future research direction for scaling BayesPO to larger optimization sets.

\section{Conclusion}

We presented \method, a Bayesian prompt optimization framework based on parallel-tempered gradient-guided discrete MCMC. The method defines a posterior over prompt tokens, constructs an energy from task likelihood and an LLM prior, uses MH-corrected GwL proposals for token-level updates, adapts the proposal to non-weight-tied LLM embeddings, and integrates PT for global exploration. Experiments show that the method can discover meaningful prompts, escape a local optimum in a controlled text completion task, and improve average accuracy on APE instruction induction when used as a post-optimizer. At the same time, the study reveals that energy minimization may overfit small optimization sets and that the current sampler remains computationally expensive. Future work will therefore focus on faster-mixing Monte Carlo methods, prefix-prediction-aligned LLM training signals, trans-dimensional prompt-length sampling, and discrete mini-batch MCMC with valid correction mechanisms.

\appendix

\section{Experiment Details and Use of Language Models}
\label{app:experiment-details}

\subsection{Reproducibility Details}
The experiments use Qwen2.5-0.5B-Instruct for diagnostic prompt optimization and Qwen2.5-7B-Instruct for PT and APE benchmark experiments. The APE experiments use 24 instruction-induction subtasks, six training examples per task, eight PT chains with temperatures $\{1.8,1.5,1.4,1.33,1.28,1.2,1.1,1.0\}$, 1000 iterations, and step size $\alpha=4.0$. Prompt length is 20 tokens except for Sentence Similarity, where it is 50. The poetry completion experiment uses nine chains with temperatures $\{2.0,1.6,1.33,1.28,1.22,1.17,1.11,1.06,1.0\}$ and 4000 iterations.

\subsection{Use of Language Models}
The authors used large language models to polish and revise the writing of the manuscript. The models were not used to generate ideas, perform analysis, or produce original scientific content.

\subsection{Poetry Initialization Template}
\label{app:poetry-init-template}
For the poetry completion experiment, the initialization of the missing middle line is generated by Qwen2.5-7B-Instruct using the following chat-template messages. The variable \texttt{middle\_len} is set to 6, \texttt{prefix\_text} is ``下面是《静夜思》中的两句诗：'', and \texttt{suffix\_text} is ``低头思故乡。''.
\begin{quote}
\begin{CodeBlock}
{
    "role": "system",
    "content": f"你是一个文本补全助手。请在给定的前缀和后缀之间，生成恰好{middle_len}个token的合理中间内容。不要生成多余内容。"
},
{
    "role": "user",
    "content": f"前缀：{prefix_text}\n后缀：{suffix_text}\n请生成中间的{middle_len}个token："
}
\end{CodeBlock}
\end{quote}

\section{Model Inference and Diagnostic Task Details}
\label{app:model-diagnostic}

\subsection{Qwen-Instruct Chat Template}
\label{app:chat-template}
All experiments use the official Qwen-Instruct chat template to organize the input sequence. The exact format is:
\begin{quote}
\begin{verbatim}
<|im_start|>system
    (prompt sequence)<|im_end|>
<|im_start|>user
    (standard input text)<|im_end|>
<|im_start|>assistant
    (standard output text)<|im_end|>
\end{verbatim}
\end{quote}
The prompt sequence, namely the optimized $\rho$, is placed in the \texttt{system} role as a global instruction. The task input $x_i$ is placed in the \texttt{user} role. The target output $y_i$ is placed in the \texttt{assistant} role and is used to compute the likelihood term $\log P(y_i\mid \rho,x_i)$. The template is kept identical for Qwen2.5-0.5B-Instruct and Qwen2.5-7B-Instruct.

\subsection{Diagnostic Task Examples}
\label{app:diagnostic-examples}
Table~\ref{tab:diagnostic-examples} lists the four examples used in each diagnostic task.

\begin{table}[t!]
\centering
\small
\begin{tabular}{p{0.47\textwidth}p{0.47\textwidth}}
\toprule
Classical Chinese translation & Antonym generation \\
\midrule
(``Hello, how are you?'', ``汝安乎？'' [Are you well?])\newline
(``Thank you very much.'', ``多谢阁下。'' [Many thanks.])\newline
(``Good morning.'', ``晨安。'' [Morning peace.])\newline
(``The cat sat on the mat.'', ``猫踞于席上。'' [The cat sits on the mat.])
&
(``热'' [hot], ``冷'' [cold])\newline
(``高兴'' [happy], ``难过'' [sad])\newline
(``上'' [up], ``下'' [down])\newline
(``同意'' [agree], ``拒绝'' [refuse]) \\
\bottomrule
\end{tabular}
\caption{Input-output examples for the two diagnostic tasks.}
\label{tab:diagnostic-examples}
\end{table}

\subsection{Diagnostic Hyperparameters}
\label{app:diagnostic-hparams}
Because the diagnostic experiment is mainly qualitative and Qwen2.5-0.5B-Instruct is small with a relatively simple energy landscape, single-chain GwL is sufficient for effective optimization. We therefore do not introduce parallel tempering in these two tasks; instead, we run ten random seeds and select the best result. To test the sampler on different sequence lengths, the prompt length is fixed to $N=10$ for the classical Chinese task and $N=20$ for the antonym task. The step size is $\alpha=4.0$ in both tasks. The fixed prefix is ``指令：'' (``Instruction:''), and these prefix tokens are not changed during sampling; they are used to encourage the LLM to output an instruction-like prompt. The total iteration count is $S=1000$ for classical Chinese translation and $S=500$ for antonym generation.

\section{APE Benchmark: Full Prompt Changes and Energy Trajectories}
\label{app:ape-details}

\subsection{Full Before/After Prompt Comparison}
\label{app:full-prompts}
Table~\ref{tab:full-prompt-changes} gives the before/after prompt texts and test accuracies for all 24 APE instruction induction subtasks. The ``APE initialization'' column corresponds to the best prompt obtained by reproducing APE; the ``Optimized prompt'' column corresponds to the lowest-energy prompt obtained by initializing BayesPO with the APE prompt and then applying GwL+PT.

\begin{footnotesize}
\setlength\LTleft{0pt}
\setlength\LTright{0pt}
\setlength{\tabcolsep}{2pt}
\begin{longtable}{@{}p{0.14\textwidth}p{0.30\textwidth}p{0.065\textwidth}p{0.30\textwidth}p{0.065\textwidth}@{}}
\caption{APE instruction induction full before/after prompt comparison.}\label{tab:full-prompt-changes}\\
\toprule
Task & APE initialization & Acc. & BayesPO prompt & Acc. \\
\midrule
\endfirsthead
\multicolumn{5}{c}{Table~\thetable{} -- continued} \\
\toprule
Task & APE initialization & Acc. & BayesPO prompt & Acc. \\
\midrule
\endhead
\midrule
\multicolumn{5}{r}{Continued on next page} \\
\endfoot
\bottomrule
\endlastfoot
Passivization & The instruction was to transform the sentence from active voice to passive voice, but it seems that my friend did not follow the & 1.00 & The instruction was to transform the sentences from active voice to passive voice, and it seems like the response did not follow the & 1.00 \\
Antonyms & The instruction was to transform the word into its opposite. However, I am wondering if there is another way to interpret this & 0.76 & The instruction was to translate a sentence into its opposite. However, I'm glad that there's no instruction to do that & 0.75 \\
Cause Selection & The instruction was to select the sentence that is more likely to be the cause of the other. However, the output does & 0.88 & The instruction was to give the sentence that was most likely to be the cause of the another. Here is the pair: & 0.92 \\
Common Concept & The instruction was to provide an output that describes the input in one word. Can you help me refine the instruction to achieve & 0.1042 & The instruction was to find a sequence that predictions a target in one word, which, help me continue the paragraph as [' & 0.1054 \\
Difference & The instruction was to take the two inputs and produce an output, where the output is the first number minus the second number & 1.00 & The instruction was to take the two numbers and produce an output, where the output is the first number minus the second number & 1.00 \\
First Letter & The instruction was to take the first letter of the word. Is that correct? Yes, based on the input-output pairs & 1.00 & The instruction was to take the first letter of each word, is this instruction unclear or incorrect based on the input/output provided & 0.98 \\
Formality & The instruction was to convert the input into an equivalent output using more formal language. However, I feel that the outputs are & 0.4156 & The instruction was to rewrite the following in an acceptable subset of mainstream formal English: apparently, they, and one investigator suggests & 0.5196 \\
Number to Word & The instruction was to write the number in words. What is the instruction that produced these input-output pairs? The instruction that & 0.99 & The instruction was to translate the following in words. Here are the instruction that gave these input-output pairs. The input is & 0.94 \\
Larger Animal & The instruction was to take two animals and return the one that is more intelligent. Is there any error in the output? & 0.51 & The instruction was to take two numbers and find the one that is more significant. I need an answer of the form \textbackslash{} & 0.66 \\
List Letters & The instruction was to write the letters of the word in uppercase and separate each letter with a space. What would be the & 0.76 & The instruction was to write the characters of the phrase in order and separate each character with a space. What would be the & 0.79 \\
Negation & The instruction was to reverse the truth value, but it seems my friend is not only reversing the truth value but also changing & 0.68 & The instruction was to reverse the truth value, but it seems the response provided not only reverse the truth value but also negate & 0.75 \\
Starting With & The instruction was to take an input sentence and output only the words that begin with the specified letter. However, there are & 0.33 & The instruction was to take an selected SMS and produce only the words which start with the given alphabet. Also add few spaces & 0.14 \\
Rhymes & The instruction was to reverse the word and remove the first letter. Is this correct? Yes, based on the provided input & 0.10 & The instruction was to continue the word by changing the previous one. Is that correct? Yes, based on your instruction, & 0.16 \\
Second Letter & The instruction was to extract the second letter of the word, but I actually wanted them to extract the penultimate (second & 0.10 & The instruction was to generate the second letter of a word, but in doing so I also output the penultimate letter of & 0.63 \\
Sentiment Analysis & The instruction was to produce the output sentiment (positive or negative) based on the given input. Can you produce some additional & 0.91 & The instruction was to return the correct sentiment (positive: negative) based on the given sentence. This movie was a complete & 0.88 \\
Pluralization & The instruction was to pluralize the word. Is there a way to test if the output is correct? Yes, you & 0.90 & ``The instruction was to pluralize the words. Is there a way to confirm that the task was completed? Yes, there & 0.99 \\
Sentence Similarity & The instruction was to generate the output based on the similarity between the two sentences, with the following possible outputs: 1 - definitely not 2 - possibly 3 - probably 4 - almost perfectly Can you please provide a new pair of sentences that would produce an output & 0.07 & The instruction was to produce an output based on the similarity between 2 sentences, with the following possible outcomes: 1 - probably not 2 - possibly 3 - probably 4 - almost perfectly Can you please generate a similar set of responses that would indicate the following & 0.00 \\
Sum & The instruction was to take two numbers as inputs and produce their sum as output. Is there a more concise way to describe & 0.95 & The instruction was to take two halves of input and make their sum into output. Is there a more concise way to say & 0.99 \\
Translation English-German & The instruction was to translate the words into German. Is it possible to produce the following output based on the given instruction? & 0.79 & The instruction was to write the text in German. Is it possible to provide the German text based on the following input? & 0.81 \\
Synonyms & The instruction was to replace the given word with its synonym, but I made an exception for certain words. Can you figure & 0.12 & The instruction was to replace the given text with its shorter alternative, I have an exception for proper nouns. I also & 0.11 \\
Membership & The instruction was to select words that were either an animal or could be an animal. However, it seems like there might & 0.81 & The instruction was to write words that are sorted and are or could be an animal (4-5 characters in a sentence & 0.49 \\
Translation English-Spanish & The instruction was to translate words from English to Spanish. Is this correct? Yes, based on the input-output pairs you & 0.66 & The instruction was to translate from English to Spanish A possible and more accurate version based on the context provided below: & 0.82 \\
Translation English-French & The instruction was to translate words from French to English. Can you explain what went wrong in the translations? The translations provided & 0.49 & The instruction was to translate words from English to French. Can you explain what went wrong in the translations? The translations received & 0.40 \\
Word in Context & The instruction was to determine if the two sentences are using the word in the same way or not, and output same & 0.08 & The instruction was to \textbackslash{} tell if the two constructions are using the same r the same construction or not, simply type same & 0.34 \\
\end{longtable}
\end{footnotesize}

\subsection{Additional APE Energy Trajectories}
\label{app:ape-energy-trajectories}
\begin{figure}[t!]
\centering
\begin{minipage}{0.49\textwidth}
\centering
\includegraphics[width=\linewidth]{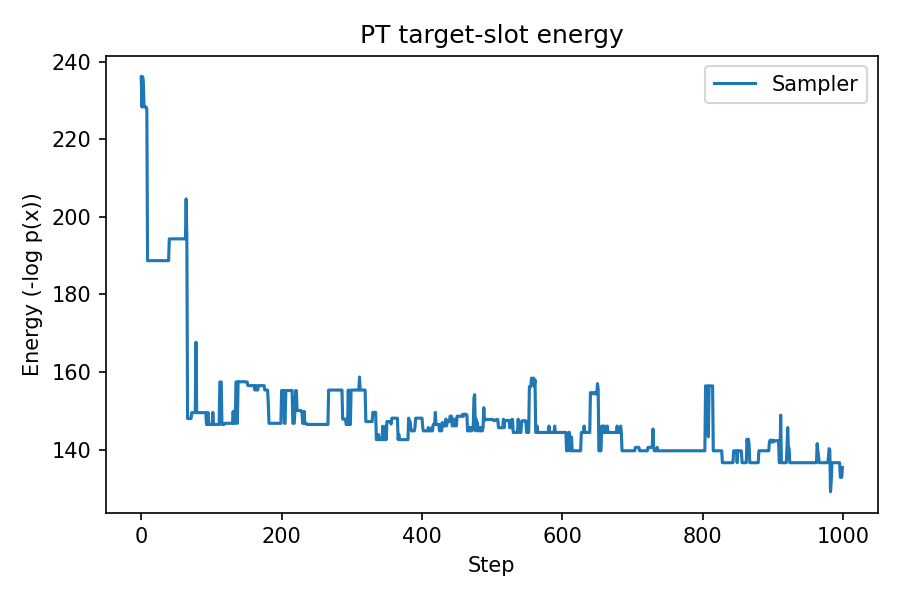}
\end{minipage}\hfill
\begin{minipage}{0.49\textwidth}
\centering
\includegraphics[width=\linewidth]{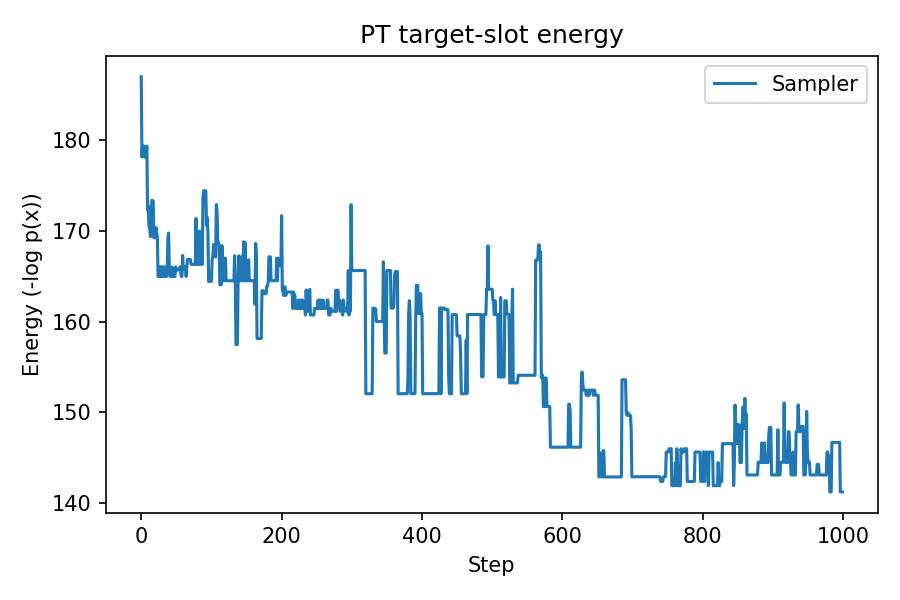}
\end{minipage}
\vspace{0.5em}
\begin{minipage}{0.49\textwidth}
\centering
\includegraphics[width=\linewidth]{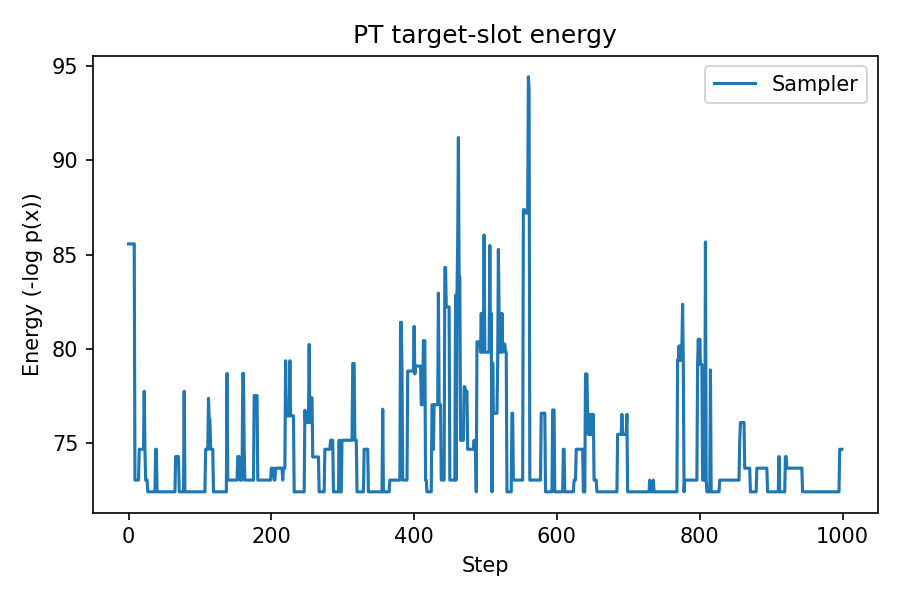}
\end{minipage}\hfill
\begin{minipage}{0.49\textwidth}
\centering
\includegraphics[width=\linewidth]{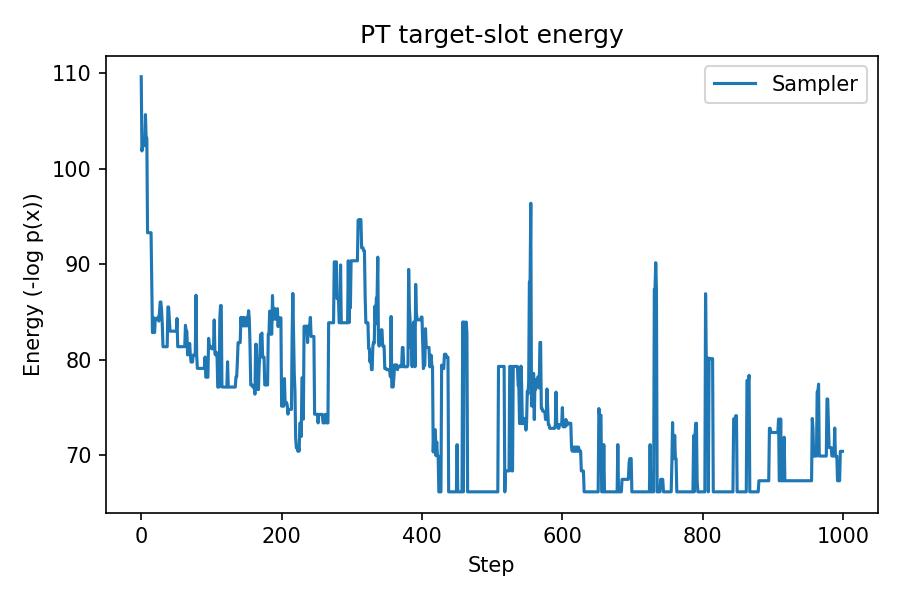}
\end{minipage}
\caption{Additional target-chain energy trajectories. Top-left: Word in Context (+26.00 points). Top-right: Starting With (-19.00 points). Bottom-left: Difference (+0.00 points). Bottom-right: Passivization (+0.00 points).}
\label{fig:appendix-energy}
\end{figure}

Figure~\ref{fig:appendix-energy} reports additional APE-task energy trajectories for tasks with different accuracy outcomes. The corresponding accuracy changes are Word in Context: +26.00 percentage points, Starting With: -19.00 points, Difference: +0.00 points, and Passivization: +0.00 points. These curves are useful diagnostics: the sampler consistently decreases the specified training energy, but the relationship between training energy and test accuracy depends on whether the optimization examples are representative.

\FloatBarrier

\bibliographystyle{iclr2026_conference}
\bibliography{bayespo_refs}

\end{CJK*}
\end{document}